\pdfoutput=1

\documentclass[11pt]{article}

\usepackage[]{ACL2023}
\usepackage{times}
\usepackage{latexsym}

\usepackage[T1]{fontenc}

\usepackage[utf8]{inputenc}

\usepackage{microtype}

\usepackage{inconsolata}
\usepackage{subfigure}
\usepackage{graphicx}
\usepackage{enumitem}
\usepackage{diagbox}
\usepackage{tabularx}
\usepackage{color}
\usepackage{booktabs}
\usepackage{amsmath}
\usepackage{adjustbox} 
\setenumerate[1]{itemsep=0pt,partopsep=0pt,parsep=\parskip,topsep=0pt}
\setitemize[1]{itemsep=0pt,partopsep=0pt,parsep=\parskip,topsep=0pt}
\setdescription{itemsep=0pt,partopsep=0pt,parsep=\parskip,topsep=0pt}
\title{Evaluate What You Can't Evaluate: Unassessable Quality for Generated Response}

\author{Yongkang Liu\textsuperscript{\rm 1,2,3}, Shi Feng\textsuperscript{\rm 1}, Daling Wang\textsuperscript{\rm 1}, Yifei Zhang\textsuperscript{\rm 1}, Hinrich Schütze\textsuperscript{\rm 2,3} \\
\textsuperscript{\rm 1} Northeastern University, China\\
\textsuperscript{\rm 2} Center for Information and Language Processing, LMU Munich\\
\textsuperscript{\rm 3} Munich Center for Machine Learning (MCML), LMU Munich\\
\texttt{misonsky@163.com, \{fengshi,wangdaling,zhangyifei\}@cse.neu.edu.cn}\\
}

\begin{document}
\maketitle
\begin{abstract}
LLMs (large language models) like ChatGPT have demonstrated exceptional language comprehension and generation abilities. While reference-free evaluators grounded in LLMs exhibit superior
human alignment compared to traditional reference-based evaluators, the utilization of such evaluators poses several challenges. Reference-free evaluators are better suited for open-ended
examples with different possible responses, but not all examples are open-ended. For closed-ended examples with unique correct semantic response, reference-free evaluators may still 
consider it high quality, even if the given response contradicts the facts and semantics of dialogue history. To provide a comprehensive assessment of the reliability of evaluators based on LLMs, we have created two adversarial meta-evaluation dialogue generation datasets: KdConv-ADV, derived from KdConv, and DSTC7-ADV, derived from DSTC7-AVSD. Compared to previous meta-evaluation benchmarks, both KdConv-ADV and DSTC7-ADV present greater challenges since they contain lots of closed-ended examples and adversarial instances derived from references. Experimental results reveal that reference-free evaluators based on LLMs are a reliable alternative to reference-based evaluators on tasks that do not involve external knowledge. Reference-free evaluators tend to overestimate the quality of the text and are still deficient in distinguishing text quality.
\end{abstract}
\section{Introduction}
The evaluation of generated response quality using reference-based metrics has faced criticism from researchers~\cite{liu2016not}. The primary reason behind this criticism stems from the fact that reference-based evaluation metrics, such as BLEU~\cite{papineni2002bleu}, ROUGE~\cite{lin2004rouge}, and METEOR~\cite{banerjee2005meteor} consider candidates with high similarity with reference responses as indication of high quality, which contradicts the semantic and expression diversity present in the responses. Therefore, reference-based metrics fail to fairly evaluate different reasonable responses, leading to a low correlation with human judgments~\cite{liu2016not,sedoc2019chateval,liu2023gpteval}.
\begin{figure}[h]
\centering
\includegraphics[width=\linewidth]{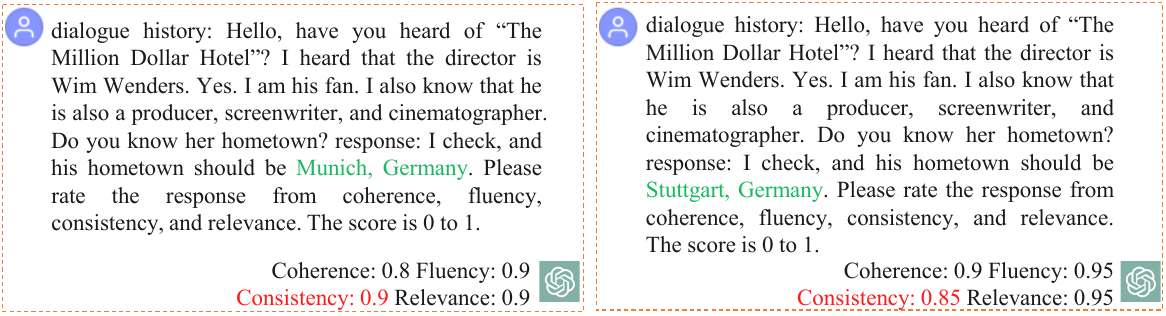}
\caption{Evaluation examples of ChatGPT. The correct response semantic for this example is unique. The reference response is \textit{I checked and his hometown should be \textcolor{blue}{Düsseldorf, Germany}}.}
\label{intro:example}
\end{figure}

Given the remarkable language understanding and generation capabilities demonstrated by LLMs~\cite{kocon2023chatgpt,frieder2023mathematical,huang2023chatgpt,qin2023chatgpt,rao2023can} like ChatGPT~\cite{ouyang2022training}, LLaMA~\cite{touvron2023llama}, and GPT-4~\cite{OpenAI2023GPT4TR}, recent studies have suggested leveraging these models as reference-free evaluators for assessing the quality of generated text~\cite{fu2023gptscore,wang2023chatgpt,liu2023gpteval}. Different from reference-based evaluators, reference-free evaluators employ LLMs to score the generated responses according to different instructions without any reference target, which can address the problem of reference-based evaluators using the reference as the sole criterion.

Although researches~\cite{fu2023gptscore,wang2023chatgpt,liu2023gpteval} show that reference-free evaluators demonstrate better human agreement, 
their reliability is questionable.
The primary reason is that previous studies~\cite{fu2023gptscore,wang2023chatgpt,liu2023gpteval} have not conducted a comprehensive evaluation of reference-free evaluators. As shown in Table~\ref{tab:sta}, the benchmark datasets Topical-Chat~\cite{gopalakrishnan2019topical} and Persona-Chat~\cite{zhang2018personalizing} utilized in existing works predominantly consist of open-ended examples with different semantic responses, lacking evaluations for closed-ended examples with unique correct semantic responses. The conversational context of open-ended examples is unrestricted, and the semantics of the corresponding responses are broad, or even arbitrary. The results only on open-ended examples do not truly reflect the accuracy and objectivity of evaluators.
The broad semantic space of responses results in many candidates being equally reasonable. It is unexplored whether evaluators have the ability to distinguish nuances in the quality of responses. As long as the evaluator gives high scores or similar scores in most cases, it will be considered reliable.
Closed-ended examples differ from open-ended examples in that their conversation context is semantically restricted, either derived from external knowledge or the dialogue history. The limitation makes the semantics of responses unique, which makes evaluators able to provide reasonable judgments only when it correctly understands the underlying limitation. Therefore, closed-ended examples can better reflect the quality of evaluators than open-ended examples.

A closed-ended example is provided in Figure~\ref{intro:example}, where \textit{"the hometown of Wim Wenders is Düsseldorf, Germany"} represents the sole accurate candidate semantic for this closed-ended instance.
Despite two unreasonable responses (i.e., \textit{"... Munich, Germany"} and \textit{"... Stuttgart, Germany"}) that are inconsistent with the fact (i.e., \textit{the hometown of Wim Wenders is Düsseldorf, Germany}), ChatGPT still gave a high score in terms of consistency dimension (i.e., 0.9 and 0.85).
When evaluating solely on open-ended examples, evaluators with significant high scoring biases may be erroneously perceived as exhibiting stronger agreement with humans, owing to the expansive semantic possibilities within the candidate space of open-ended examples.

To address these challenges, we build two adversarial meta-evaluation dialogue generation datasets KdConv-ADV and DSTC7-ADV based on KdConv~\cite{zhou2020kdconv} and DSTC7-AVSD~\cite{alamri2019audio}, respectively. Meta-evaluation is a process that assesses the quality of evaluation methods.
In contrast to prior meta-evaluation dialogue datasets~\cite{mehri2020usr}, both KdConv-ADV and DSTC7-ADV encompass not only open-ended examples but also lots of closed-ended examples. Specifically, the KdConv-ADV consists of equal numbers of closed-ended and open-ended examples. We ask human annotators to generate three new candidate responses with low lexical overlap with the reference response for each example. The generated candidates demonstrate both reasonability and high quality for open-ended examples, while the generated candidates tend to be inconsistent with the provided facts, and even include fictitious information (i.e., adversarial examples) for closed-ended instances.
The DSTC7-ADV is completely composed of closed-ended examples. We generate adversarial examples by rewriting the reference responses to ensure that their semantics are inconsistent with the provided facts. 
Candidate responses that have a low lexical overlap with the reference in KdConv-ADV may be of high-quality, whereas candidates that have a high overlap with the reference in DSTC7-ADV may be of low-quality, making reference-based metrics almost useless, which are also extremely challenging for reference-free evaluators.

We evaluate ChatGPT and multiple open source LLMs, such as Vicuna~\cite{vicuna2023} and ChatGLM~\cite{du2022glm}. Experimental results on KdConv-ADV and DSTC7-ADV show that reference-free evaluators based on LLMs have the following disadvantages: i) insufficient knowledge; ii) insufficient ability to identify unreasonable responses; iii) insufficient differentiation of scores.
To summarize, we make the following contributions:
\begin{itemize}[leftmargin=*]
    \item We construct two adversarial meta-evaluation dialogue datasets \textbf{KdConv-ADV} and \textbf{DSTC7-ADV} based on KdConv and DSTC7-AVSD to comprehensively evaluate the reliability of dialogue generation metrics.
    \item We propose new challenges for dialogue generation metric evaluation, requiring evaluators to be able to evaluate generated text at a semantic level rather than lexical matching.
    \item We evaluate and analyze the performance of reference-based and reference-free evaluators on \textbf{KdConv-ADV} and \textbf{DSTC7-ADV}. Experimental results show that LLM-based reference-free evaluators demonstrate promising performance as alternatives to reference-based methods, particularly for tasks not requiring external knowledge.
\end{itemize}
\section{RELATED WORK}
\subsection{Reference-based Evaluators}
\paragraph{Ngram-based Metrics}
Ngram-based metrics evaluate the dialogue models by measuring the lexical overlap between a generated response and a reference text. BLEU~\cite{papineni2002bleu}, METEOR~\cite{banerjee2005meteor}, ROUGE~\cite{lin2004rouge} are widely used metrics for dialogue generation evaluation~\cite{bao2020plato,liu2022mulzdg,liu2022pvgru}. Most of these metrics are based on n-gram overlap between a generated candidate and reference response. They fail to measure the content quality of generated candidates and therefore do not evaluate the dialogue generation systems accurately. ~\citet{honovich2021q2} proposes to use a question answering system for fact consistency evaluation. This method relies on high-quality external knowledge and question answering system. ~\citet{dziri2022evaluating} introduces a new benchmark to evaluate the reliability of reference-based metrics.

\paragraph{Embedding-based Metrics}
Embedding-based Metrics evaluate the dialogue generation systems by measuring the semantic similarity between the generated candidate and the reference response. Embedding Average is a metric that measures the distance between two texts by averaging the vector representations of their constituent words, which is widely used in textual similarity tasks~\cite{wieting2015TowardsUP,liu2016not,liu2022mulzdg}. BERTScore~\cite{Zhang2019BERTScoreET} employs the contextualized representation from BERT to measure the similarity between generated candidate and reference response. MoverScore~\cite{zhao2019moverscore} adds soft alignments based on BERTScore to obtain a more robust similarity measurement. ~\citet{ghazarian2020predictive} trains a classification task on pooled vectors to evaluate the engagement of responses. 

These methods pay more attention to semantic similarity than ngram-based metrics but still fail to make a fair assessment for multiple reasonable responses because embedding-based metrics still consider candidates with high overlap with the reference to be of high quality.
\subsection{Reference-free Evaluators}
Reference-free evaluation refers to methods of judging the quality of generated text according to the degree of correlation between dialogue history and generated candidates in multiple aspects. Existing works usually trained specific models as reference-free evaluators before LLMs. ~\citet{mehri2020usr} 
proposes an unsupervised reference-free metric by training models on downstream tasks to evaluate open-ended examples. ~\citet{pang2020towards} uses data augmentation methods to train a more robust evaluators. ~\citet{yeh2021comprehensive} shows that metrics that rely on a specific data set lack the ability to generalize. ~\citet{dziri2022faithdial} create FAITHDIAL datasets based on the Wizard of Wikipedia to reduce factual errors in the training corpus. ~\citet{khalid2022explaining} proposes an adversarial test-suite to evaluate the bias of metrics based on trained models.
However, reference-free evaluation by humans is still a must in almost all dialogue generation tasks~\cite{zhang2018context,zhou2020kdconv,bao2020plato,liu2022mulzdg}.
Human evaluation is expensive and only evaluates a small number of examples that are selected. 
Reference-free evaluators based on LLMs offer hope for solving this problem. 

~\citet{wang2023chatgpt} believes that ChatGPT achieves competitive correlation with golden human judgments through preliminary meta-evaluation. ~\citet{liu2023gpteval} proposes the probability of each score calculated by the ChatGPT or GPT-4~\cite{OpenAI2023GPT4TR} as the weight for the corresponding score to improve the alignment with human judgment. GPTScore~\cite{fu2023gptscore} takes the sum of the logarithms of the decoding probabilities of the evaluated text as the final score. According to existing experimental conclusions, evaluators based on LLMs have a tendency to replace human evaluaiton. However, these studies lack evaluation on closed-ended examples as well as stability of evaluators testing. Therefore, we construct two adversarial meta-evaluation dialogue datasets to test the reliability of evaluators on closed-ended  and adversarial examples. Different from the traditional works~\citep{honovich2021q2,dziri2022evaluating,khalid2022explaining} of evaluation reference-based and model-based metrics, we pay more attention to the reliability of reference-free evaluators based on LLMs.
\section{Dataset Construction}
For the existing meta-evaluation datasets (i.e., Topical-Chat and Persona-Chat) constructed by ~\citet{mehri2020usr}, we manually annotate the types (i.e., open-ended and closed-ended) of examples. 
The standard for annotation is to mark as closed-ended if the response to be generated does not have semantic diversity according to the dialogue history and facts provided, otherwise it is open-ended. The labels are annotated based on the dialogue history when no facts. 
We construct two new adversarial meta-evaluation datasets KdConv-ADV and DSTC7-ADV that include lots of closed-ended examples. 
The statistical results of the datasets are shown in Table~\ref{tab:sta} (Appendix~\ref{app:results}). We can observe that the datasets we built contains a large number of closed-ended instances with adversarial examples, which can test the reliability of different evaluators on closed-ended examples and the stability on adversarial examples. Datasets KdConv-ADV and DSTC7-ADV are derived from KdConv and DSTC7-AVSD, respectively.
\subsection{KdConv-ADV}
KdConv is a Chinese multi-domain Knowledge-driven Conversation dataset~\cite{zhou2020kdconv}, which provides a reference response for each example. We select 91 examples with unique response semantics
from KdConv as closed-ended examples. The characteristic of these examples is to use unique information such as location or time as the response. For example, \textit{"Tokyo is the capital of Japan."}
We pick an equal number of instances with multiple response semantics from KdConv as open-ended examples. The questions in these examples are open-ended, and the semantics of the responses are not unique, such as \textit{"What do you think of Tokyo?"} 
In order to effectively evaluate whether evaluators have the ability to identify unreasonable responses under low lexical overlap, we ask annotators to generate three new candidate responses for each example.

For closed-ended examples, the generated candidate responses are \textbf{inconsistent} with dialogue histories and are even \textbf{false information}. To achieve this goal, we utilize GPT-4 to generate five candidates according dialogue history. The specific prompt is \textit{"Dialogue history: \$content\$. Please generate five different responses"}, where \$content\$ represents the content of the conversation history. For the five generated responses, we intentionally alter key candidate information to be irrelevant or even false. Finally, we select three candidates with the lowest BLEU scores compared to the reference response as test examples. As shown in Table~\ref{tab:case}, the location information (i.e., \textit{"Taipei, Taiwan"}) provided by the first candidate is inconsistent with fact, and the locations in the second and third candidates are completely fictitious (i.e, \textit{"Yamaguchi Prefecture, in Chang'an Kyushu"} and \textit{Matsuyama City, Ehime Prefecture, Tokyo}).

For open-ended examples, we expect the generated candidates to be reasonable. Similarly, we utilize GPT-4 to generate five candidates according dialogue history. Then we then manually perform information correction on the generated candidates based on the original knowledge base provided by KdConv. We also select the three candidates with the lowest BLEU scores compared to the reference response as test examples. As shown in Table~\ref{tab:case}, the generated candidates are of \textbf{high quality} and \textbf{reasonable}.

\begin{figure*}[t]
\centering 
\includegraphics[width=0.85\textwidth]{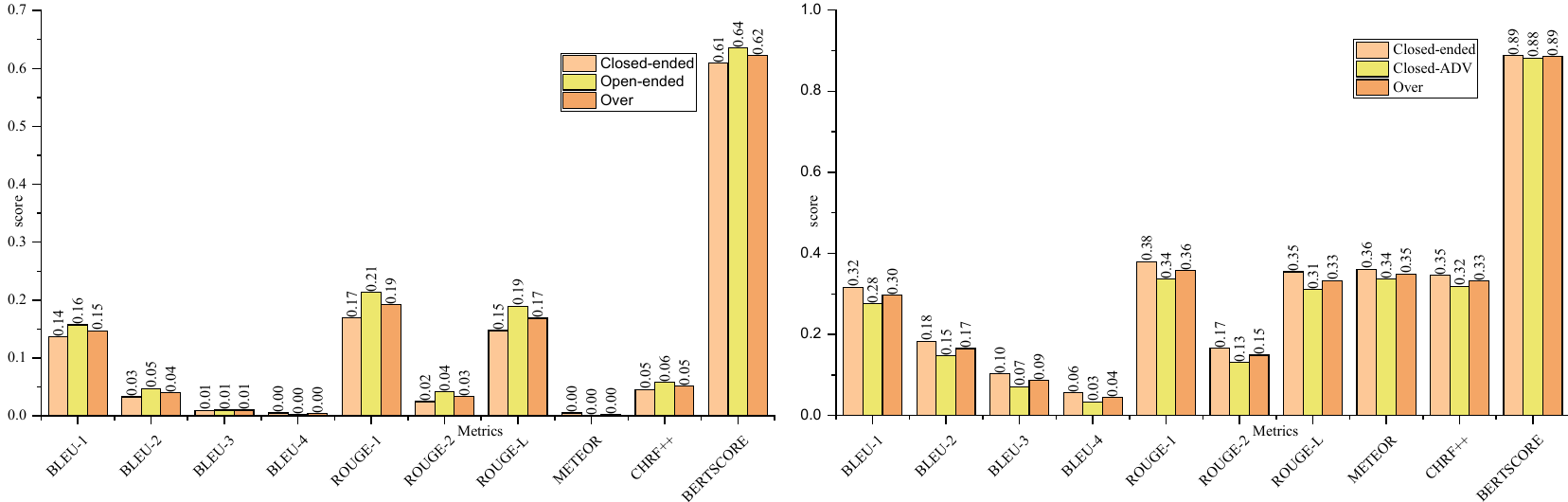}
\caption{The distribution of reference-based metrics for different types of examples on KdConv-ADV (left) and DSTC7-ADV (right). The \textbf{over} corresponds to the performance of overall datasets.}
\label{refer_dis}
\end{figure*}

Using the responses provided by KdConv as references to score candidates based on different metrics. As shown in Figure~\ref{refer_dis} (left), we can observe that the BLEU-1 score of the closed-ended examples is 14\%, and that of the open-ended examples is 16\%, which means the lexical overlap is low between generated candidate and reference responses in KdConv-ADV. For open-ended examples, although there is low lexical overlap between candidate and reference responses, these candidates are high-quality and reasonable responses.
It can be found that these reference-based metrics give almost similar scores to closed-ended and open-ended examples, which indicates that reference-based metrics does not have the ability to identify unreasonable responses when the reference and candidate responses have low lexical overlap.

Notably, we refrain from leveraging the knowledge provided by the corpus during response evaluation. This approach stems from our desire to assess the agreement between various evaluators and human judgment without relying on external knowledge, which is made due to the challenge of furnishing accurate knowledge bases for each example in most cases.
\subsection{DSTC7-ADV}
DSTC7-AVSD is a knowledge-grounded response generation dataset with textual knowledge that is video’s caption and summary~\cite{alamri2019audio}. DSTC7-AVSD provides six reference candidates with similar semantics and different expressions. We consider the first one as the reference response, while the remaining ones are regarded as candidate responses.
We select 342 examples with unique response semantics from DSTC7-AVSD as closed-ended examples. In order to effectively evaluate whether evaluators are able to identify unreasonable responses based on semantics rather than matching, we reverse the semantics of responses to obtain the same amount of adversarial examples by negation transformations, such as \textit{"can"} $\rightarrow$ \textit{"can not"}, \textit{"is"} $\rightarrow$ \textit{"is not"}, \textit{"only"} $\rightarrow$ \textit{"not only"}.
Therefore, candidate response semantics of adversarial examples are contradictory to the facts provided (i.e., video descriptions). As shown in Table~\ref{tab:case}, the responses of the adversarial examples (i.e., \textit{"...not only one..."}, \textit{"...not only a single...",\textit{"...two persons..."} and \textit{"...seven persons..."}}) are contradictory and inconsistent with the facts (i.e., "...one person..."). We analyze the characteristics of DSTC7-ADV by calculating the scores of reference-based evaluators. As shown in Figure~\ref{refer_dis} (right), we can observe that the BLEU-1 score of the closed-ended examples is 32\%, and that of the corresponding adversarial examples (i.e., closed-ADV) is 28\%, which means the lexical overlap is higher between generated candidate and reference responses compared to KdConv-ADV. Similar phenomena can also be observed from the results of other metrics. For adversarial examples, although there is high lexical overlap between candidate and reference responses, these candidates are unreasonable responses.
Besides, we can also find that these metrics give almost similar scores to closed-ended and adversarial examples, which shows that these metrics do not have the ability to identify unreasonable responses based on semantics.

Different from KdConv-ADV, we employ facts provided when evaluating responses. We follow previous studies~\cite{bao2020plato,liu2022pvgru} to concatenate video descriptions to conversation history beginnings. The primary motivation is to assess the proficiency of various evaluators in comprehending and utilizing knowledge effectively.
\paragraph{Evaluation Dimensions}
Based on previous studies~\cite{zhang2018context,bao2020plato,xu2022long,liu2022pvgru}, we divide the reference-free evaluation dimensions into two categories: \textbf{independent} and \textbf{correlated} dimensions. Independent dimensions are evaluated solely based on the generated candidates, without considering any other factors or references, mainly including \textbf{fluency}, \textbf{naturalness} and \textbf{engagingness}. The correlated dimensions refers to the evaluation not only based on candidates but also referring to the relationship between candidates and dialogue history even facts, mainly including \textbf{coherence}, \textbf{relevance}, \textbf{consistency} and \textbf{groundedness}.

Existing studies~\cite{mehri2020usr,liu2023gpteval} based on Topical-Chat and Persona-Chat have extensively studied evaluators on four dimensions: \textit{naturalness}, \textit{coherence}, \textit{engagingness} and \textit{groundedness}. 
We select \textit{coherence}, \textit{relevance}, \textit{consistency} and \textit{fluency} not tested before as evaluation dimensions for DSTC7-ADV and KdConv-ADV, and find that evaluators based on LLMs are more likely to make mistakes in correlated evaluation dimensions after preliminary experimental analysis. The definition of each evaluation dimension is defined as follows:
\begin{itemize}[leftmargin=*]
    \item \textbf{Fluency} refers to the fluency and grammatical correctness of responses.
    \item \textbf{Coherence} refers to the logical and semantic coherence between responses and previous context.
    \item \textbf{Relevance} refers to the degree to response is connected or relevant to a particular topic, question, or situation of previous context.
    \item \textbf{Consistency} refers to the logical and factual consistency between responses and previous context, facts also include external commonsense knowledge.
\end{itemize}
\begin{table*}[ht]
\begin{adjustbox}{max width=0.85\textwidth, center}
\includegraphics[width=\textwidth]{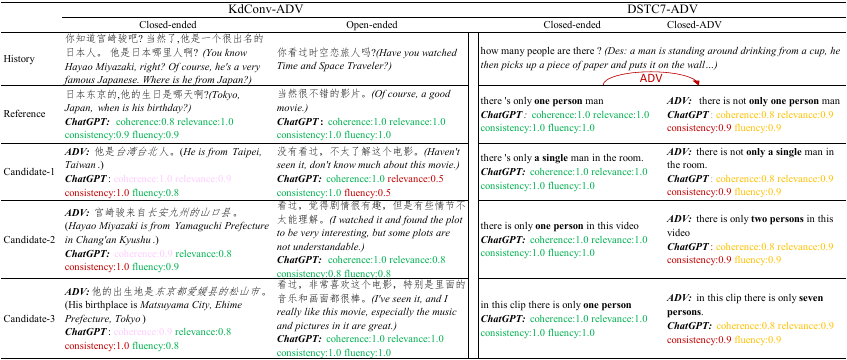}
\end{adjustbox}
\caption{Examples of KDConv-ADV and DSTC7-ADV. \textbf{ADV} indicates that the corresponding candidate is an adversarial example. The score corresponding to \textcolor{green}{green} indicates that the evaluation is reasonable, \textcolor{pink}{pink} indicates a slightly higher score, \textcolor{red}{red} indicates that the evaluation is unreasonable, and \textcolor{brown!70!yellow}{yellow} indicates unreasonable fluctuations in the scores of adversarial examples.}
\label{tab:case}
\end{table*}
\paragraph{Prompt for Evaluation}
Note that the reference-free evaluator is a prompt-based evaluation process. We find that the graded scoring mechanism may lead to the low variance of the scores and the low correlation with human judgments~\cite{liu2023gpteval}. Another fact is that the ranking mechanism (i.e., ordering of candidates during scoring) will get different results due to the different positions of multiple candidates~\cite{wang2023large}. In order to compare with the with reference-based evaluators, we divide each dimension into 10 levels (i.e., L1:(0-0.1), L2(0.1-0.2),...,L10(0.9-1)) (Appendix~\ref{app:evaluation}). We find that the output may focus on one value when asking LLMs to output a level number, such as 10, making it impossible to calculate Spearman~\cite{zar2005spearman} and Spearman coefficientscitep~\cite{mukaka2012guide}. In order to avoid this problem, we require LLMs to output a value of 0-1, which is mapped to the corresponding level finally.
We follow previous studies~\cite{huang2023chatgpt,liu2023gpteval} to design the prompts, as shown in Figure~\ref{fig:prompt} (Appendix~\ref{sec:PromptTemplate}).
Note that different LLMs correspond to different delimiters. If there is no fact, the content of the corresponding position is empty. In this manner, the dialogue history, response, corresponding fact and the definition of evaluation dimension are given to LLMs. Next, LLMs will give its judgment (e.g., "\textit{The response is consistent with the information provided in the input. Therefore, the score is 1.}"). Finally, the numerical scores could be easily extracted via heuristic rules. 
To evaluate the consistency between LLMs and human judgement, we performed human annotation.
There are three annotators for human annotation, the average of the three points is used as the final score. The final score will be mapped to the corresponding level. The Fleiss' Kappa~\citep{moons2023measuring} is 0.766, which indicates better annotation agreement. Please refer to the appendix for details (i.e., Appendix~\ref{app:evaluation}).
\begin{figure*}
  \centering
  \includegraphics[width=0.85\textwidth]{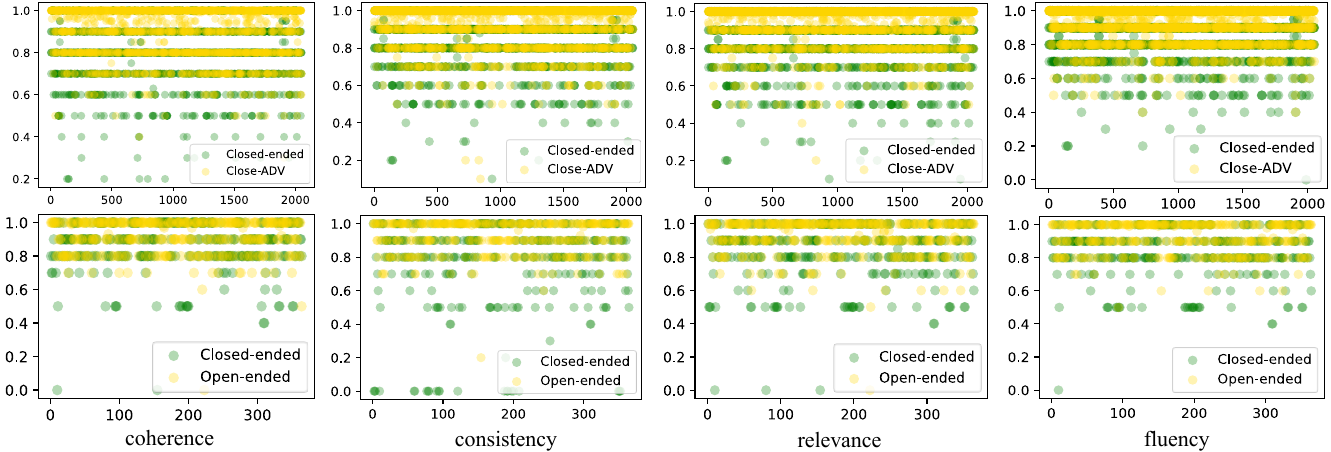}
\caption{The score distribution of ChatGPT on DSTC7-ADV (up) and KDConv-ADV (down).}
\label{chatgpt_score}
\end{figure*}
\section{Experiments}
The introduction of the baselines~\ref{app:base} and detailed experimental setup~\ref{app:exp} are in the Appendix.
\begin{table*}[ht]
\begin{adjustbox}{max width=0.98\textwidth, center}
\includegraphics[width=\textwidth]{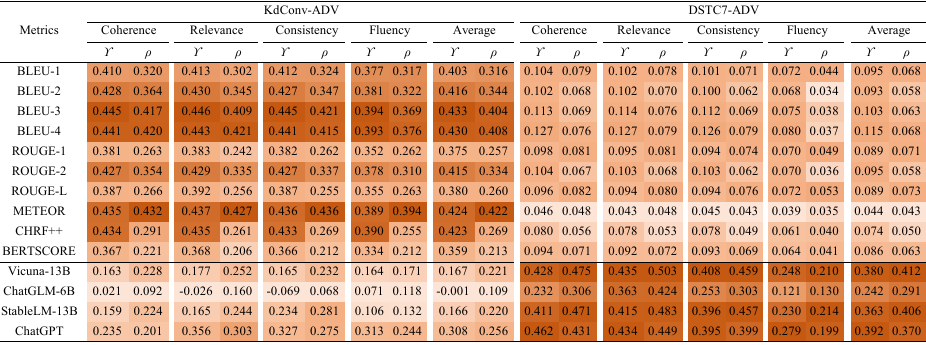}
\end{adjustbox}
\caption{Turn-level Spearman ($\rho$) and Pearson ($\gamma$) correlations of different metrics on KdConv-ADV and DSTC7-ADV.}
\label{tab:origin_table}
\end{table*}

\begin{table*}[ht]

\begin{adjustbox}{max width=0.98\textwidth, center}
\includegraphics[width=\textwidth]{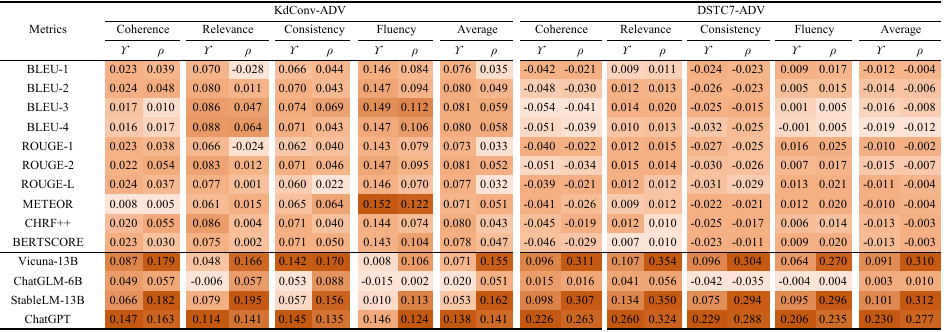}
\end{adjustbox}
\caption{Turn-level Spearman ($\rho$) and Pearson ($\gamma$) correlations of different metrics on open-ended examples.}
\label{tab:open_table}
\end{table*}
\subsection{Results}
To test the agreement between different evaluators and human on the dialogue response generation task, we compute turn-level Pearson and Spearman correlation on Topical-Chat, Persona-Chat, KdConv-ADV and DSTC7-ADV. Table~\ref{tab:topical-per} (Appendix~\ref{app:results}) reports the results of different evaluators on Topical-Chat and Persona-Chat. We can observe that reference-free evaluators have better human agreement compared to reference-based evaluators. Specifically, the spearman correlation of UNIEVAL evaluator on Topical-Chat is 53.3\%, and pearson correlation is 57.7\%. The ChatGPT evaluator achieve similar results to UNIEVAL on Topical-Chat. ChatGPT evaluator's spearman is 45\% and pearson is 39\% on Persona-Chat. On KdConv-ADV and DSTC7-ADV datasets, we can draw the same conclusion from Table~\ref{tab:open_table} that reference-free evaluators have better human agreement. 

Table~\ref{tab:origin_table} reports the results of different evaluators on KdConv-ADV and DSTC7-ADV.
Traditional reference-based evaluators have better human agreement compared to reference-free evaluators based on LLMs on KdConv-ADV. The results of BLEU-3, BLEU-4 and METEOR outperform evaluator based on ChatGPT by an average of 12.5\%/14.8\% (pearson/spearman), 12.2\%/15.2\% and 11.6\%/16.6\% respectively.
However, we observe the opposite result where reference-free evaluators based on LLMs outperform reference-based evaluators on DSTC7-ADV. Evaluator based on ChatGPT outperforms BLEU-4 by an average of 27.7\% and 30.2\%. The disparate phenomena observed in the two datasets suggest that reference-free evaluators encounter reliability issues. The reasons for this phenomenon are complex. We will conduct an in-depth analysis from the perspective of datasets and evaluators.
\begin{table*}[ht]

\begin{adjustbox}{max width=\textwidth, center}
\includegraphics[width=0.95\textwidth]{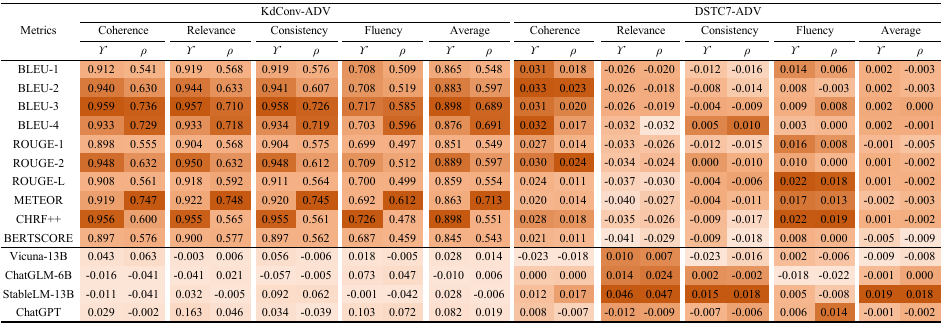}
\end{adjustbox}
\caption{Turn-level Spearman ($\rho$) and Pearson ($\gamma$) correlations of different metrics on closed-ended examples.}
\label{tab:close_open}
\end{table*}
\subsection{Reliability of Reference-based Evaluators}
To further analyze the performance of different evaluators on different data types, we report results of evaluators on different data types separately.
Table~\ref{tab:open_table} and Table~\ref{tab:close_open} report the performance of different evaluators on open-ended and closed-ended examples (i.e., adversarial examples) on KdConv-ADV and DSTC7-ADV.
An interesting phenomenon is that reference-based evaluators show better alignment with humans in KdConv-ADV's closed-ended examples, which leads to reference-based evaluators having better alignment on KdConv-ADV. On DSTC7-ADV dataset, the results are completely opposite. 
The main reason is that the candidates and references of KdConv-ADV's closed-ended examples have low overlap (i.e., Figure~\ref{refer_dis}), which causes reference-based evaluators that judge text quality based on lexical matching to tend to give low scores. And the candidates of KdConv-ADV's closed-ended examples that are inconsistent with history are of low quality. The tendency to give low scores and the consistency of low-quality responses are important reasons for high alignment between reference-based evaluators and humans. However, a high overlap between candidates and responses does not mean that the candidates are of high quality on DSTC7-ADV, which is the causes the reference-based evaluators to fail.
We believe that using reference-based evaluators may result in unfair evaluation for tasks that generate text with high diversity, such as dialogue generation tasks. But for tasks with low diversity, such as translation tasks, extractive generation tasks, etc., reference-based evaluators are still a more credible choice.
\subsection{Reliability of Reference-free Evaluators} Different from reference-based evaluators, reference-free evaluators use LLMs to score the generated responses without any reference target. 
As we observed, the reference-free evaluators have better alignment with humans on open-ended examples (i.e., Table~\ref{tab:open_table} and Table~\ref{tab:topical-per}). For evaluation tasks involving knowledge, the reference-free evaluators will give unfair judgement without sufficient external knowledge support. According to Table~\ref{tab:origin_table}, reference-free evaluators have poorer alignment with humans compared to reference-based evaluators on KdConv-ADV. We also observe that the reference-free evaluators achieves better human alignment on DSTC7-ADV when external knowledge is provided. We consider reference-free evaluators to be a reliable alternative to reference-based evaluators on tasks that do not involve external knowledge. 

Most tasks involve external knowledge, which requires LLMs to be a knowledgeable evaluators to make a reasonable judgment. However, LLMs cannot update its knowledge in real time. Therefore, how to make full use of external knowledge bases and improve the reliability of reference-free evaluators for knowledge-based task evaluation is challenging.
\paragraph{Discrimination Ability} An effective evaluator should demonstrate the ability to identify unreasonable responses and distinguish responses of varying qualities. To reveal whether LLMs have the ability to distinguish responses of different quality, we take ChatGPT as an example to report its score distribution on KdConv-ADV and DSTC7-ADV, as shown in Figure~\ref{chatgpt_score}. The scores of KdConv-ADV are mostly concentrated between 0.8 and 1.0 and the scores of DSTC7-ADV are mostly concentrated between 0.5 and 1.0. Although the score distribution of ChatGPT in DSTC7-ADV is more discriminative than that in KdConv-ADV, the scores of ChatGPT still have the tendency of overestimation on DSTC7-ADV (i.e., most of the scores are above 0.5). While it is evident that the responses from the adversarial example of DSTC7-ADV contradict the factual information, it is surprising that lots of adversarial examples achieve a high score of 0.9 on the consistency dimension, which is extremely unreasonable.

There are still some deficiencies in reference-free evaluators based on LLMs. First, LLMs have inherent limitations in their knowledge. Second, the scores of LLMs have a large room for improvement in distinguishing responses of different qualities. The ratings of LLMs exhibit a tendency to cluster within a narrow range, displaying low variance, and sometimes even assigning high ratings to unreasonable responses.
\section{Conclusion}
We construct two adversarial meta-evaluation dialogue datasets KdConv-ADV and DSTC7-ADV. Based on KdConv-ADV and DSTC7-ADV, we analyze the performance and reliability of reference-free and reference-based evaluators. We think that reference-based evaluators are still reliable for tasks with low diversity, and reference-free evaluators are a reliable alternative to reference-based evaluators on tasks that do not involve external knowledge. Reference-free evaluators may provide unreasonable evaluations for tasks involving knowledge when external knowledge is absent. Besides, reference-free evaluators tend to overestimate the quality of the text and are still deficient in distinguishing text quality.
\section*{Limitations}
While we analyze the challenges and possibilities and of LLMs as text generation evaluators by constructed benchmarks, the utilization of LLMs as evaluators for text generation is in the exploratory phase. There are limitations that provide avenues for future work: i) the performance of LLMs as an NLG metric is related to prompts, how to reduce the sensitivity of LLMs to prompt and improve the reproducibility of results is an important issue.
ii) our work pays more attention to dialogue tasks with high diversity, and lacks the analysis of other generative tasks.
\bibliography{custom}
\bibliographystyle{acl_natbib}

\appendix
\section{Appendix}
\label{sec:appendix}

\subsection{Tables}
\label{app:results}
\begin{table}[ht]
\begin{adjustbox}{max width=\columnwidth, center}
\includegraphics[width=\textwidth]{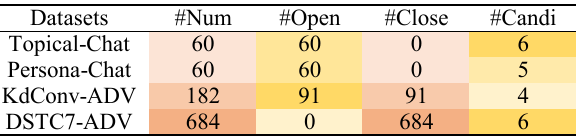}
\end{adjustbox}
\caption{ \textbf{Dataset statistics.} \textit{\#Num} represents the total number of examples. \textit{\#Open} represents the number of open-ended examples. \textit{\#Close} represents the number of closed-ended examples, including a large number of adversarial examples. \textit{\#Candi} represents the number of candidate responses. Meta-evaluation datasets Topical-Chat and Persona-Chat can be downloaded from \url{http://shikib.com/usr}.}
\label{tab:sta}
\end{table}

\begin{table*}[!t]

\begin{adjustbox}{max width=0.95\textwidth, center}
\includegraphics[width=\textwidth]{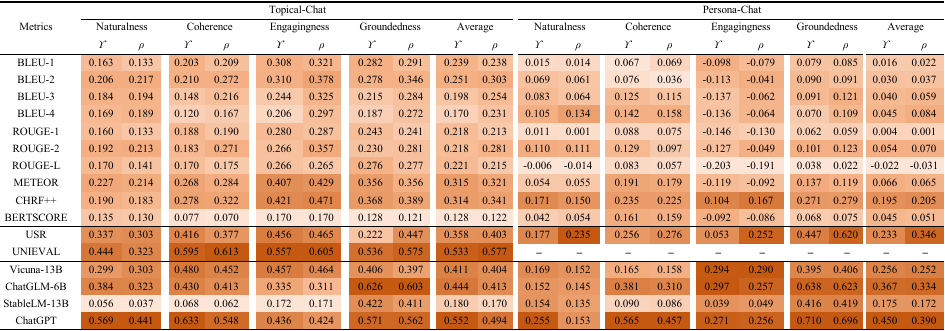}
\end{adjustbox}
\caption{Turn-level Spearman ($\rho$) and Pearson ($\gamma$) correlations of different metrics on Topical-Chat and Persona-Chat.}
\label{tab:topical-per}
\end{table*}

\subsection{Prompt Template}
\label{sec:PromptTemplate}
\begin{figure}[h]
\centering
\includegraphics[width=0.90\linewidth]{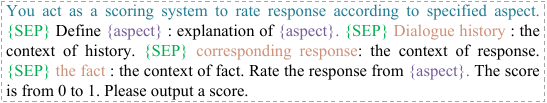}
\caption{Prompt Template. The \textbf{\{aspect\}} denotes the evaluation dimension, such as fluency. The explanation of \textbf{\{aspect\}} includes corresponding level definitions. The \textbf{\{SEP\}} represents the delimiter.}
\label{fig:prompt}
\end{figure}
\subsection{Case Study}
\label{sec:case}
In order to analyze the existing problems more intuitively, we present detailed cases (i.e., Table~\ref{tab:case}). It can be clearly seen that ChatGPT gives high scores for adversarial examples on KDConv-ADV and DSTC7-ADV. In the example of KdConv-ADV, ChatGPT fails to recognize fictional locations (i.e., \textit{"Yamaguchi Prefecture in Chang'an Kyushu"} and \textit{"Matsuyama City, Ehime Prefecture, Tokyo"}) and gives the highest rating on the consistency dimension.
In the example of DSTC7-ADV, ChatGPT cannot identify candidates with semantic inconsistencies with given fact caused by slight perturbations. As mentioned in previous subsection, considerable text comprehension abilities is the premise and basis for using LLMs as evaluators. However, we can conclude that the robustness of LLMs has a lot of room for improvement, and LLMs cannot correctly understand subtle semantic perturbations.
\subsection{Evaluation Dimensions}
\label{app:evaluation}
We select \textbf{coherence}, \textbf{relevance}, \textbf{consistency} and \textbf{fluency} as evaluation dimensions. The level definitions for different dimensions are as follows.

\textbf{Fluency}:
\begin{itemize}[leftmargin=*]
    \item L1: almost incomprehensible, heavily grammatical errors, poor coherence.
    \item L2: There are many grammatical errors, and it is difficult to understand.
    \item L3: Many grammatical errors, unclear expressions, require effort to understand.
    \item L4: There are some grammatical errors, and the expression is acceptable.
    \item L5: There are some grammatical errors, and the expression is generally clear.
    \item L6: The grammar is basically correct, and the expression is coherent, but there are some minor errors.
    \item L7: The grammar is correct, the expression is fluent and coherent, and there are only a few minor errors.
    \item L8: The grammar is almost completely correct, and the expression is fluent and natural, with some minor errors.
    \item L9: The grammar is almost perfect, the expression is very fluent, and there are few errors.
    \item L10: Whether it is grammar, vocabulary, or expression, it is perfect, with almost no errors.
\end{itemize}

\textbf{Consistency}:
\begin{itemize}[leftmargin=*]
    \item L1: lack logical structure and coherence, contain internal inconsistencies and fake information, making it difficult to follow or understand.
    \item L2: contain internal inconsistencies, where statements within contradict each other.
    \item L3: contradict the established context, either within the conversation history.
    \item L4: contain factual inaccuracies or incorrect information that can be easily identified based on available knowledge.
    \item L5: align with the conversation history but may deviate from established facts or external knowledge.
    \item L6: demonstrate basic logical consistency within the conversation history, but there may be minor inconsistencies.
    \item L7: align well with the conversation history, with some errors or inconsistencies.
    \item L8: demonstrate logical coherence, and statements align with each other and the conversation history.
    \item L9: not only logically consistent but also reflect factual information, aligning well with the conversation history.
    \item L10: not only internally consistent and factually accurate but also align with external commonsense knowledge.
\end{itemize}

\textbf{Relevance}:
\begin{itemize}[leftmargin=*]
    \item L1: no connection to the topic of conversation.
    \item L2: touch upon the topic but lack a substantial connection.
    \item L3: contain some relevant elements but miss key points.
    \item L4: contain most relevant elements but lack a comprehensive or coherent connection to the entire context.
    \item L5: show a moderate degree of relevance, involving some aspects of conversation topic. 
    \item L6: generally relevant to the topic or question, providing a basic understanding for conversation.
    \item L7: demonstrate a good level of relevance, involving most aspects of conversation topic.
    \item L8: highly relevant, involving the topic and providing a detailed content connection to the context.
    \item L9: exhibit great level of relevance, thoroughly involving the conversation topic.
    \item L10: not only completely relevant but also involving the topic with clarity and insight.
\end{itemize}

\textbf{Coherence}:
\begin{itemize}[leftmargin=*]
    \item L1: lack both logical and semantic coherence, making it challenging to understand.
    \item L2: have semantic gaps or disjointed elements, making it difficult to establish a connection to previous context.
    \item L3: lack logical structure, leading to difficulties in understanding the conversation. 
    \item L4: exhibit partial coherence but still contain logical or semantic gaps.
    \item L5: have a basic logical relation, but there are gaps or inconsistencies in semantic coherence. 
    \item L6: demonstrate a basic level of both logical and semantic coherence, providing a generally understandable ideas.
    \item L7: exhibit a good level of logical and semantic coherence, making it easy to understand and follow.
    \item L8: display a better level of both logical and semantic coherence, ensuring a smooth and connection to previous context.
    \item L9: show an high level of both logical and semantic coherence, with a seamless and clear connection to previous context.
    \item L10: achieve flawless logical and semantic coherence, presenting information in a way that is not only easy to follow but also understanding.
\end{itemize}
Note that these are also reference standards for human annotation. We have three annotators for each dimension, and the average of the three points is used as the final score, and the final score is the result after retaining two decimal places. All annotators are graduate students engaged in NLP research. The fees incurred for labeling are supported by the corresponding funds.
\subsection{Baselines}
\label{app:base}
The reference-based evaluators used are as follows:
\begin{itemize}[leftmargin=*]
    \item \textbf{BLEU-1}, \textbf{BLEU-2}, \textbf{BLEU-3}, \textbf{BLEU-4}~\cite{papineni2002bleu}, \textbf{ROUGE-1}, \textbf{ROUGE-2} and \textbf{ROUGE-L}~\cite{lin2004rouge} measure the lexical overlap between the generated text and the candidate text.
    \item \textbf{METEOR}~\cite{banerjee2005meteor} calculates the similarity between the candidate text and the reference text based on word-level precision and recall, as well as penalties for word order.
    \item \textbf{ChrF++}~\cite{popovic2017chrf++} uses the F-score statistic for character n-gram matches to judge the similarity between the candidate text and the generated text.
    \item \textbf{BERTScore}~\cite{Zhang2019BERTScoreET} evaluates the semantic similarity via pre-trained BERT model.
\end{itemize}
The reference-free evaluators used are as follows:
\begin{itemize}[leftmargin=*]
    \item \textbf{ChatGPT} is an advanced AI language model developed by OpenAI, trained on a vast amount of text data, capable of understanding and generating human-like text across a wide range of topics.
    \item \textbf{Vicuna-13B} is an open-source chatbot trained by fine-tuning LLaMA-13B on user-shared conversations collected from ShareGPT.
    \item \textbf{ChatGLM-6B} is an open-source dialogue language model based on General Language Model~\cite{du2022glm} that supports both Chinese and English.
    \item \textbf{StableLM-13B} is an open-source dialogue language model based on vicuna fine-tuned by RLHF~\cite{ouyang2022training}.
\end{itemize}
\subsection{Experimental Setup}
\label{app:exp}
For ChatGPT (i.e., GPT-3.5), we obtain the result by calling the API interface of OpenAI\footnote{https://chat.openai.com}. We set parameters temperature to 0.7, the presence penalty to 0, the frequency penalty to 0.2 and the maximum sentence length to 1024. Codes for other LLMs are available online\footnote{https://github.com/lm-sys/FastChat}. The maximum decoding length is set to 512 for Vicuna-13B~\cite{vicuna2023}, ChatGLM-6B~\cite{du2022glm,zeng2022glm}, StableLM-13B and Dolly-12B. The temperature is set to 0.8 for Vicuna-13B and others are set to 0.7. Except for ChatGPT, the weights of other LLMs can be downloaded from the hugging face\footnote{https://huggingface.co/models}. We employ the hugging face evaluation library~\footnote{https://huggingface.co/evaluate-metric} to calculate the results for reference-based metrics. The default delimiter is "\#\#\#". For ChatGPT and Vicuna we use space as delimiter. We use RTX A6000 (48G) for inference of open source LLMs.
Note that we cost about \$20 to call the ChatGPT API interface.
\subsection{Metrics}
\label{app:metrics}
We employ \textit{Spearman correlation}~\cite{zar2005spearman} ($\rho$) and \textit{Pearson correlation}~\cite{mukaka2012guide} ($\gamma$) to evaluate different metrics correlate with human judgment.

\end{document}